\documentclass{article}

\usepackage[nonatbib,preprint]{neurips_2020}
\usepackage[utf8]{inputenc} 
\usepackage[T1]{fontenc}    
\usepackage{hyperref}       
\usepackage{url}            
\usepackage{booktabs}       
\usepackage{amsmath,amssymb,amsthm}
\usepackage{nicefrac}       
\usepackage{microtype}      
\usepackage[inline]{enumitem}
\usepackage[table,usenames,dvipsnames]{xcolor}
\usepackage{graphicx}
\usepackage{tikz}
\usetikzlibrary{calc,fit,arrows}
\usetikzlibrary{positioning,intersections}
\usetikzlibrary{matrix,chains}
\usetikzlibrary{decorations.pathreplacing}
\usepackage{bm}
\usepackage{mathtools}
\usepackage{xspace}
\usepackage{caption}
\usepackage{subcaption}
\usepackage{wrapfig}
\usepackage[linesnumbered,ruled,noline]{algorithm2e}
\usepackage{comment}
\let\en=\ensuremath

\newtheorem{definition}{Definition}

\newtheorem{lemma}{Lemma}

\allowdisplaybreaks

\DeclarePairedDelimiter{\norm}{\lVert}{\rVert}

\DeclarePairedDelimiterX{\divx}[2]{(}{)}{#1\;\delimsize\|\;#2}

\newcommand{\pdv}[2]{\en{\frac{\partial #1}{\partial #2}}}

\DeclareMathOperator{\E}{\mathbb{E}}          
\DeclareMathOperator{\Divergence}{\mathfrak{D}}

\newcommand{\Div}[0]{\Divergence_{\alpha}\divx}

\DeclareMathOperator{\Range}{range}

\renewcommand{\vec}[1]{\en{\bm{\mathrm{#1}}}}

\newcommand{\mat}[1]{\en{{\bm{\mathrm{#1}}}}}

\newcommand{\R}[0]{\mathbb{R}}

\renewcommand{\th}[0]{\textsuperscript{th}\xspace}

\graphicspath{{./figures/}}

\title{Differentially Private Deep Learning with Direct Feedback Alignment}

\author{%
  Jaewoo Lee\\
  University of Georgia\\
  Athens, GA 30602 \\
  \texttt{jwlee@cs.uga.edu} \\
  \And
  Daniel Kifer \\
  The Pennsylvania State University \\
  University Park, PA \\
  \texttt{dkifer@cse.psu.edu} \\
}

\allowdisplaybreaks

\begin{document}

\maketitle

\begin{abstract}
Standard methods for differentially private training of deep neural networks replace back-propagated mini-batch gradients with biased and noisy approximations to the gradient. These modifications to training often result in a privacy-preserving model that is significantly less accurate than its non-private counterpart.

We hypothesize that alternative training algorithms may be more amenable to differential privacy. Specifically, we examine the suitability of direct feedback alignment (DFA). We propose the first differentially private method for training deep neural networks with DFA and show that it achieves significant gains in accuracy (often by 10-20\%) compared to  backprop-based differentially private training on a variety of architectures (fully connected, convolutional) and datasets.

%
%
%
\end{abstract}



\section{Introduction}
\label{sec:introduction}
An unanswered question, with significant implications, is what is the best way to train deep networks with differential privacy.
The non-private setting has seen rapid advances in the state-of-the-art while progress in the privacy-preserving setting has been lagging.
Currently there are two promising privacy-preserving approaches, each with its own drawbacks. (1) Knowledge distillation approaches, such as PATE \cite{pate,Papernot2018pate}, requires massive quantities of \emph{public} data (i.e., data that will \emph{not} receive privacy protections) in addition to massive amounts of sensitive data (which will be protected). These requirements limit their applicability. (2) Adaptations of stochastic gradient descent \cite{Abadi2016deep,Yu2019modelpub,AcsGAN,thakkarclip,Brendan2018learning,ChenDPGAN,BeaulieuJones159756,Abay2018:PPS}, are more widely applicable but result in low accuracy compared to non-private models. They operate by separately clipping the gradient for each example in a batch before aggregating --- that is, they clip then aggregate, instead of using the aggregate-then-clip approach common in non-private training \cite{Goodfellow-et-al-2016}. Noise is added to the gradient estimate and then parameters are updated. The result is a \emph{biased} and noisy gradient that causes final model accuracy to deteriorate \cite{bagdasaryan2019differential}.

In this paper, we focus on the setting where the entire dataset is sensitive and we seek alternatives to differentially private stochastic gradient descent. Specifically, we consider the suitability of direct feedback alignment (DFA) \cite{nokland2016direct} for privacy-preserving training of artificial neural networks. DFA is a biologically inspired alternative to backprop/SGD that is much more suitable for low-power hardware  \cite{Han_2019_ICCV,crafton2019}. We propose the first differentially private algorithm for DFA. A careful analysis shows privacy can be achieved by (i) clipping the activations and error signal after (not during) the feed-forward phase, (ii) carefully choosing the error transport matrix, and (iii) adding Gaussian noise to the update direction.

Empirical results on a variety of datasets show the following results. (1) Differentially private DFA outperforms (by a wide margin) differentially private SGD on fully connected networks with various activation functions. (2) Networks with fully connected layers stacked on top of convolutional layers benefit significantly from a hybrid approach that combines DFA at the top layers with SGD at the bottom layers (mirroring an earlier result in the non-private setting \cite{Han_2019_ICCV}).

\section{Related Work}
\label{sec:related_work}
Backpropagation (BP)~\cite{rumelhart1986learning} and 
stochastic gradient descent (SGD) have been essential 
tools for training modern deep learning models in both non-private and
private settings. In some communities, such as cognitive neuroscience, debates about 
 biological implausibility of BP (e.g., weight transport
problem~\cite{grossberg1987competitive,liao2016important,akrout2019deep}) motivated
 approaches
such as feedback alignment (FA)~\cite{lillicrap2016random}, direct
feedback alignment (DFA)~\cite{nokland2016direct}, and difference target
propagation (DTP)~\cite{lee2015difference}, which introduce new learning
mechanisms for error feedback signals. 
Recent work~\cite{bartunov2018assessing,moskovitz2018feedback,crafton2019}
found that biologically plausible learning algorithms
generally underperform BP on convolutional neural networks, but hybrid DFA/BP approaches can work well \cite{Han_2019_ICCV}.

The first differentially private training algorithm for deep learning models
was proposed by Shokri and Shmatikov~\cite{RezaDeep}.
It relied on a distributed system in which participants jointly
train a model by exchanging perturbed SGD updates. However, it incurred a privacy cost
that was too large to be practical (e.g., $\epsilon$ values in the
hundreds or thousands).
Abadi et al.~\cite{Abadi2016deep} proposed a training algorithm that satisfied Renyi Differential Privacy \cite{Mironov2017renyi}, which allowed training of networks with more practical privacy costs (epsilon values in the single digits). It modified SGD by individually clipping the gradient of each example before aggregating them into batches, then adding appropriately-scaled Gaussian noise, and then updating parameters as in standard SGD. This per-example-gradient clipping was designed to control the influence of any single data point. They also introduced a \emph{moment accountant} to keep track of privacy protections due to Gaussian noise and subsampling (in the selection of batches).
Their method has generated substantial interest, and follow-up
research~\cite{Yu2019modelpub,AcsGAN,thakkarclip,Brendan2018learning,Mcmahan2018general,ChenDPGAN,BeaulieuJones159756,Abay2018:PPS}
investigated additional architectures and efficient clipping
strategies. Clipping gradients before aggregating them in batches is
known to produce bias --- the estimated mini-batch gradient no longer
points in the same direction as the true mini-batch gradient. As we
show experimentally, this bias already causes a degradation in model
accuracy (even without noise addition).


In some special cases, gradient clipping can be avoided. 
Phan et al.~\cite{PhanAAAI2016} focused on learning auto-encoders by
perturbing the objective function. Xie et al.~\cite{Xie:DPGAN} showed
that a differentially private Wasserstein GAN \cite{pmlr-v70-arjovsky17a} can be trained by clipping weights instead of gradients.
PATE~\cite{pate,Papernot2018pate} also avoids gradient clipping, but requires a large \emph{non}-sensitive
dataset in order to operate.

To the best of our knowledge, we propose the first differentially private version of direct feedback alignment.




\section{Preliminaries}
\label{sec:prelim}
We next introduce our notation and provide background on DFA and (Renyi) Differential privacy.

\subsection{Notation}
We use bold-face uppercase letters (e.g., $\mat{W}$) to represent
matrices, bold-face lowercase letters (e.g., $\mathbf{x}$) to represent
vectors and non-bold lowercase letters (e.g., $y$) to represent scalars. Tensors of order 3 or
higher (i.e., multidimensional arrays  indexed by 3 or more
variables) are represented in calligraphic font
(e.g,. $\mathcal{W}$). 
We index vectors using square brackets (e.g., $\mathbf{x}[1]$ is the
first component of the vector $\mathbf{x}$). For matrices, we use
subscripts to identify entries ($W_{i,j}$ is the entry in row $i$,
column $j$). 

We use the following notation to represent a dataset $D$ and its constituent records:  $D = \{\vec{d}_i = (\vec{x}_i, \vec{y}_i)\}_{i=1}^n$ is a set of $n$
examples, where $\textbf{x}_i$ is a feature vector (e.g., image pixels) and $\vec{y}_i$ is
a one-hot encoded target vector (i.e., to represent class $k$, $\vec{y}_i[k]=1$ and all other components are 0).

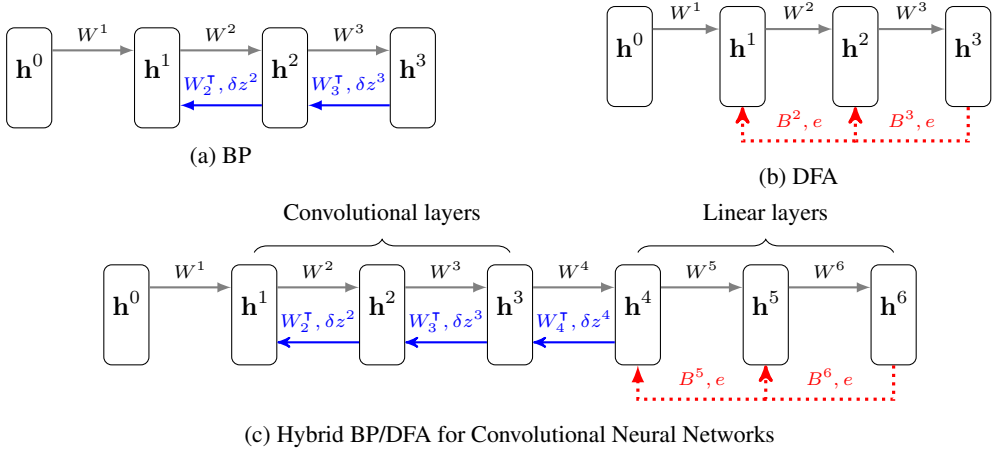
\begin{figure}[tp]
  \centering
  \begin{subfigure}[c]{.45\textwidth}
    \centering
    \begin{tikzpicture}[%
      mynode/.style={circle,draw=none,fill=none,minimum width=1em,minimum height=1em,on chain},
      mylayer/.pic={%
        \begin{scope}[start chain=going below, node distance=0pt]
          \foreach \i in {1,...,3}{%
            \node[mynode] (-x\i) {};
          }
          \node[fit={(-x1)(-x3)},draw,rounded corners=3pt] (-bb) {#1};
        \end{scope}}]
      \foreach \y in {0,...,3}{%
        \pic[local bounding box=l\y] (l\y) at (\y*1.7, 0) {mylayer={$\vec{h}^{\y}$}};
      }      
      \foreach \l [evaluate=\l as \p using int(\l-1)] in {1,...,3}{%
        \draw[-latex,thick,draw=gray] (l\p-x1-|l\p-bb.east) --node[above,font=\scriptsize] {$W^{\l}$}
        (l\l-x1-|l\l-bb.west);
      }
      \foreach \l [evaluate=\l as \p using int(\l-1)] in {3,2}{%
        \draw[-latex,blue,thick,draw] (l\l-x3-|l\l-bb.west)
        --node[above,midway,font=\scriptsize]
        {$W_{\l}^\intercal, \delta z^{\l}$}
        (l\p-x3-|l\p-bb.east);
      }
    \end{tikzpicture}
    \caption{BP}
    \label{fig:BP}
  \end{subfigure}
  \hfill
  \begin{subfigure}[c]{.45\textwidth}
    \centering
    \begin{tikzpicture}[%
      mynode/.style={circle,draw=none,fill=none,minimum width=1em,minimum height=1em,on chain},
      mylayer/.pic={%
        \begin{scope}[start chain=going below, node distance=0pt]
          \foreach \i in {1,...,3}{%
            \node[mynode] (-x\i) {};
          }
          \node[fit={(-x1)(-x3)},draw,rounded corners=3pt] (-bb) {#1};
        \end{scope}}]
      \foreach \y in {0,...,3}{%
        \pic[local bounding box=l\y] (l\y) at (\y*1.5, 0) {mylayer={$\vec{h}^{\y}$}};
      }      
      \foreach \l [evaluate=\l as \p using int(\l-1)] in {1,...,3}{%
        \draw[-latex,thick,draw=gray] (l\p-x1-|l\p-bb.east) --node[above,font=\scriptsize] {$W^{\l}$}
        (l\l-x1-|l\l-bb.west);
      }
      \draw[-stealth',very thick,dotted,red] (l3-bb.south) --++ (0, -.45) node (ue){} -|
      node[above,pos=0.25,font=\scriptsize] {$B^3,e$} (l2-bb.south);
      \draw[-stealth',very thick,dotted,red] (l2-bb.south|-ue) -|
      (l1-bb.south) node[above,pos=0.25,font=\scriptsize] {$B^2,e$};
    \end{tikzpicture}
    \caption{DFA}
    \label{fig:DFA}
  \end{subfigure}
  \begin{subfigure}[b]{\textwidth}
    \centering
    \begin{tikzpicture}[%
      mynode/.style={circle,draw=none,fill=none,minimum width=1em,minimum height=1em,on chain},
      mylayer/.pic={%
        \begin{scope}[start chain=going below, node distance=0pt]
          \foreach \i in {1,...,3}{%
            \node[mynode] (-x\i) {};
          }
          \node[fit={(-x1)(-x3)},draw,rounded corners=3pt] (-bb) {#1};
        \end{scope}}]
      \foreach \y in {0,...,6}{%
        \pic[local bounding box=l\y] (l\y) at (\y*1.7, 0) {mylayer={$\vec{h}^{\y}$}};
      }      
      \foreach \l [evaluate=\l as \p using int(\l-1)] in {1,...,6}{%
        \draw[-latex,thick,draw=gray] (l\p-x1-|l\p-bb.east) --node[above,font=\scriptsize] {$W^{\l}$}
        (l\l-x1-|l\l-bb.west);
      }
      \draw[-stealth',very thick,dotted,red] (l6-bb.south) --++ (0, -.45) node (ue){} -|
      node[above,pos=0.25,font=\scriptsize] {$B^6,e$} (l5-bb.south);
      \draw[-latex,very thick,dotted,red] (l5-bb.south|-ue) -| (l4-bb.south) node[above,pos=0.25,font=\scriptsize] {$B^5,e$};
      \foreach \l [evaluate=\l as \p using int(\l-1)] in {4,3,2}{%
        \draw[-stealth',blue,thick,draw] (l\l-x3-|l\l-bb.west)
        --node[above,midway,font=\scriptsize] {$W_{\l}^\intercal, \delta z^{\l}$}
        (l\p-x3-|l\p-bb.east);
      }
      \draw[decorate,decoration={brace,amplitude=5pt,raise=3pt}]
      ([yshift=5pt]l1-x1.north) -- ([yshift=5pt]l3-x1.north)
      node[midway,above=10pt,font=\footnotesize] {Convolutional layers}; 
      \draw[decorate,decoration={brace,amplitude=5pt,raise=3pt}]
      ([yshift=5pt]l4-x1.north) -- ([yshift=5pt]l6-x1.north)
      node[midway,above=10pt,font=\footnotesize] {Linear layers}; 
    \end{tikzpicture}
    \caption{Hybrid BP/DFA for Convolutional Neural Networks}
    \label{fig:CNN_hybrid}
  \end{subfigure}  
  \caption{Comparison of different error transportation configurations}
  \label{fig:error_prop}
\end{figure}


\subsection{Backpropagation and DFA}
\label{sec:prelim_bp}
Direct Feedback Alignment \cite{nokland2016direct} is best explained by showing how it deviates from backprop. 

As an example, consider a feed-forward network $f_{\theta}$, consisting of $L$ fully-connected
layers with a soft-max output (for a classification task). Let $\mat{W}^l \in \R^{n_l\times n_{l-1}}$ and $\vec{b}^l \in \R^{n_l}$
denote the weight and bias of the $l$\th layer in $f_{\theta}$,
respectively. The feed-forward step of layer is defined by:
\begin{equation}
  \vec{h}^l= \phi_l(\vec{z}^l)\,,\quad \vec{z}^l =
  \mat{W}^l\vec{h}^{l-1} + \vec{b}^l\,,\quad \theta =\{(\mat{W}^l, \vec{b}^l)\}_{l=1}^{L}\,,
\end{equation}
where $\vec{z}^l \in \R^{n_l}$ is the pre-activation and $\vec{h}^l
\in \R^{n_l}$ is the (post-) activation of the layer obtained by applying
the activation function $\phi_l$ element-wise on
$\vec{z}^l$. In our notation, $\vec{h}^0=\vec{x}$ denotes the input
record. In the final layer (with softmax activation), 
the $k$\th output is 
$\vec{h}^L[k] = \frac{\exp(\vec{z}^L[k])}{\sum_{j} \exp(\vec{z}^L[j])}$,
which can be interpreted as the probability estimate 
for class $k$. Continuing our example, suppose the network is to be trained with
cross-entropy loss. That is, if $f_\theta(\vec{x})\equiv\hat{\vec{y}}$ is the output vector for input $\vec{x}$ with true class label $\vec{y}$, the cross-entropy loss for that record is $\mathcal{L}(\vec{h}^L) = -\sum_{k} \vec{y}[k]\log \hat{\vec{y}}[k]$.

In BP, the gradient of $\mathcal{L}$ w.r.t. the parameter $\mat{W}^l$
is computed using the chain rule and updated as follows:
\begin{align}
  \delta \vec{z}^l
  &=  \pdv{\vec{h}^l}{\vec{z}^l}\pdv{\vec{z}^{l+1}}{\vec{h}^l}\pdv{L}{\vec{z}^{l+1}}
    =
    \begin{cases}
      \left((\mat{W}^{l+1})^\intercal\delta\vec{z}^{l+1}\right)\odot\phi_l^\prime(\vec{z}^l)
      & \mbox{ if $l<L$,}\\
      \hat{\vec{y}} - \vec{y}
      &\mbox{ if $l=L$,}
    \end{cases} \label{eq:bp_error_prop}\\
  \pdv{\mathcal{L}}{\mat{W}^l}
  &= \pdv{\vec{z}^l}{\mat{W}^l}\pdv{L}{\vec{z}^{l}}   = \delta\vec{z}^l(\vec{h}^{l-1})^\intercal   \label{eq:grad_param} \\
  \mat{W}^l &\gets \mat{W}^l - \eta \pdv{\mathcal{L}}{\mat{W}^l}; \quad\vec{b}^l \gets \vec{b}^l-\eta \delta\vec{z}^l \label{eq:bpsgd} 
\end{align}
where $\eta$ is the step size, $\odot$ is an element-wise multiplication
operator and $\phi^\prime_l$ denotes the derivative of the activation function. Equation~\eqref{eq:bp_error_prop} 
shows that the backward error signal $\delta \vec{z}^l$ for layer
$l$ is computed using the signal $\delta \vec{z}^{l+1}$ propagated from
the  layer above. Starting from the output layer, the error information
propagates backward through the network from layer to layer.
Notice that the error signal $\delta \vec{z}^l$ requires the transpose
of forward weight matrix $\mat{W}^{l+1}$. In other 
words, $\mat{W}^{l+1}$ appears in both feed-forward and backward
paths. 

DFA makes two modifications to the backward path of BP (and only the backward path). First, for each layer, it 
replaces $\mat{W}^{l+1}$ in the backward path with a random feedback
weight matrix $\mat{B}^{l+1}$ (chosen at the beginning of training). The entries of $\mat{B}^{l+1}$ are randomly sampled from a
probability distribution such as the Gaussian distribution and then are
fixed throughout the training process (i.e. they do not get updated/learned). The second change introduced by DFA is that
instead of propagating the error signal backward through the layers, the error signal at each layer depends directly on $\delta\vec{z}^L$, which is the  error signal of the output layer. Due to the special importance of $\delta\vec{z}^L$, we denote this error signal  by $\vec{e}$ (which equals $\hat{\vec{y}}-\vec{y}$ for cross-entropy loss over softmax output). Mathematically, the error signal $\delta\vec{z}^l$ at layer $l$ is:
\begin{equation} \label{eq:dfa_error_prop}
  \delta\vec{z}^l = \left(\mat{B}^{l+1}\vec{e}\right) \odot \phi_l'(\vec{z}^l)\,.
\end{equation}
Compared to Equation \ref{eq:bp_error_prop}, $(\mat{W}^{l+1})^\intercal$ is replaced with $\mat{B}^{l+1}$ while $\delta\vec{z}^{l+1}$ is replaced with $\vec{e}$. Plugging Equation \ref{eq:dfa_error_prop} into Equation \ref{eq:grad_param} and then \ref{eq:bpsgd}, one arrives at the DFA update equation for $\mat{W}^l$ (the weights used in the feed-forward pass) and $\vec{b}^l$:
\[
  \mat{W}^l \gets \mat{W}^l -\eta
  \left(\left(\mat{B}^{l+1}\vec{e}\right) \odot \phi'(\vec{z}^l)\right)(\vec{h}^{l-1})^\intercal\,;\quad \vec{b}^l \gets \vec{b}^l - \eta\left(\left(\mat{B}^{l+1}\vec{e}\right) \odot \phi'(\vec{z}^l)\right)
\]

Graphically, the distinction between BP and DFA is illustrated in
Figures~\ref{fig:BP} and \ref{fig:DFA}.

\subsection{Differential Privacy}\label{subsec:dp}
Differential privacy is a formal notion of privacy that provides
strong privacy protection in sensitive data analysis. It bounds the influence that one record can have on the output of a randomized algorithm.

We say two datasets $D_1$ and $D_2$ are \emph{neighbors}
if $D_1$ can be obtained from $D_2$ by changing one
record and write $D_1 {\sim} D_2$ to denote this relationship.

\begin{definition}[($\epsilon,\delta$)-DP~\cite{Dwork2006calibrating,Dwork2006our}]
  \label{def:dp}%
Given privacy parameters $\epsilon\geq 0$, $\delta\geq 0$, a
randomized mechanism (algorithm) $\mathcal{M}$ satisfies ($\epsilon, 
\delta$)-differential privacy if for every set $S \subseteq
\Range(\mathcal{M})$ and for all pairs of neighboring datasets $D_1{\sim}D_2$, 
\[
  \Pr[\mathcal{M}(D_1) \in S] \leq \exp(\epsilon)\Pr[\mathcal{M}(D_2)\in
  S] + \delta\,.
\]%
The probability is only with respect to the randomness in $\mathcal{M}$.
\end{definition}

In differentially private deep learning, a mechanism corresponds to the set of parameter updates from processing one minibatch. Since training involves many minibatch updates, it is important to accurately track  the combined privacy leakage from all minibatches used. In this work, we use R\'enyi Differential
Privacy (RDP)~\cite{Mironov2017renyi} to track privacy leakage. RDP uses its own parameters, but after training finishes, the RDP parameters can be converted
 to the $\epsilon,\delta$ parameters of Definition \ref{def:dp}. 
RDP relies on the concept of R\'enyi divergence:
\begin{definition}[R\'enyi Divergence]
Let $P_1$ and $P_2$ be probability distributions over a set $\Omega$ 
and let $\alpha\in (1,\infty)$. R\'enyi $\alpha$-divergence $\Divergence_\alpha$
is defined as:
$\Div{P_1}{P_2} = \frac{1}{\alpha-1}\log(\E_{x\sim P_2}\left[P_1(x)^\alpha P_2(x)^{-\alpha}\right])$.
\end{definition}
R\'enyi differential privacy requires two parameters: a moment
$\alpha$ and a parameter  $\epsilon$ that bounds the moment.
\begin{definition}[$(\alpha,\epsilon)$-RDP~\cite{Mironov2017renyi}]
Given a privacy parameter $\epsilon \geq 0$ and an $\alpha\in(1,\infty)$,
a randomized mechanism $\mathcal{M}$ satisfies $(\alpha,\epsilon)$-R\'enyi
differential privacy (RDP) if for all $D_1$ and $D_2$ that differ on the
value of one record,
$
\Div{\mathcal{M}(D_1)}{\mathcal{M}(D_2)} \leq \epsilon\,.
$
\end{definition}
A simple way to achieve $(\alpha, \epsilon)$-RDP is to take a vector-valued deterministic function $f$ and add appropriately scaled Gaussian noise as follows. The scale of the noise depends on the sensitivity of $f$.

\begin{definition}[$L_2$ sensitivity]
Let $L_2$ sensitivity of $f$, denoted by $\Delta_f$ is equal to $\sup_{D_1\sim D_2} ||f(D_1)-f(D_2)||_2$, where the supremum is taken over all pairs of neighboring datasets.
\end{definition}

\begin{lemma}[Gaussian Mechanism \cite{Mironov2017renyi}]\label{lemma:gaussmech}
  Let $f$ be a function. Let $\alpha>1$ and $\epsilon>0$. Let $M$ be the mechanism that, on input $D$, returns $f(D)+N(0,\sigma^2\mat{I})$, where $\sigma^2 = \frac{\alpha\Delta^2_f}{2\epsilon}$. Then $M$ satisfies $(\alpha,\epsilon)$-RDP.
\end{lemma}

The composition theorem of RDP states that if $M_1,\dots, M_k$ are mechanisms and each $M_i$ satisfies $(\alpha,\epsilon_i)$-RDP, then their combined privacy leakage satisfies $(\alpha, \sum_i \epsilon_i)$-RDP \cite{Mironov2017renyi}.
The parameters of RDP can be converted into those of
$(\epsilon, \delta)$-DP through the following conversion result~\cite{Mironov2017renyi}. 
\begin{lemma}[Conversion to $(\epsilon,\delta)$-DP~\cite{Mironov2017renyi}] \label{pro:conversion}
  If $\mathcal{M}$ satisfies $(\alpha, \epsilon)$-RDP, it satisfies
  $(\epsilon^\prime, \delta^\prime)$-differential privacy whenever
  $\epsilon^\prime \geq \epsilon + \frac{\log (1/\delta)}{\alpha-1}$
  and $\delta^\prime \geq \delta$. 
\end{lemma}
The following lemma states that the privacy guarantee of an $(\alpha,
\epsilon)$-RDP mechanism $\mathcal{M}$ is amplified when it is applied
on poisson subsampled data.

\begin{lemma}[Subsampled Mechanism and Privacy Amplification for
  RDP~\cite{pmlr-v89-wang19b}] \label{lemma:rdp_amp} 
 For a randomized mechanism $\mathcal{M}$ and a
dataset $D$, define $\mathcal{M} \circ \mathfrak{S}_q$ as (i) sample a
subset $B\subseteq D$ (with $q=|B|/|D|$),  by
sampling without replacement (ii) apply $\mathcal{M}$ on $B$. Then if $\mathcal{M}$
satisfies $(\alpha, \epsilon(\alpha))$-RDP with respect to $B$,
$\mathcal{M}\circ \mathfrak{S}_q$ satisfies $(\alpha,
\epsilon'(\alpha))$-RDP with respect to $D$ for any integer $\alpha
\geq 2$, where 
$\epsilon'(\alpha) \leq \frac{1}{\alpha-1} \log \big(
1+q^2 {\binom{\alpha}{2}}\min\left\{4(e^{\epsilon(2)}-1), 2e^{\epsilon(2)}, e^{\epsilon(2)}(e^{\epsilon(\infty)}-1)^2\right\} +
  \sum_{l=3}^\alpha
{\binom{\alpha}{l}} q^l  e^{(l-1)\epsilon(l)}\min\{2, (e^{\epsilon(\infty)}-1)^j\}
\big)$. 
\end{lemma}



\section{Differentially Private DFA}
\label{sec:algorithm}
In this section, we propose a differentially private version of direct feedback alignment.
Given a positive constant $c$, the clipping function shrinks the norm of a vector until it is at most $c$. Formally,
$
  \mathsf{clip}_c(\vec{v}) = \min(c, \norm{\vec{v}}_2)\frac{ \vec{v}}{ \norm{\vec{v}}_2}\,.
$
Our algorithm\footnote{This algorithm is for the ``modify a record'' definition of neighbors in differential privacy. For the add/remove version, we replace fixed minibatch sizes  with Poisson sampling, use the Poisson amplification result \cite{zhu2019poission}, and add up (instead of averaging) the updates on lines \ref{line:w} and \ref{line:b} in Algorithm \ref{alg:dp-dfa}.}
for privatizing DFA has several components: (1) first, we require that the (sub)derivatives of each activation function $\phi_l$ be bounded by a constant $\gamma_l$ (i.e., $|\phi_l^\prime(v)|\leq \gamma_l$ for all scalars $v$). This is true for the most commonly used activations such as ReLU, sigmoid, tanh, etc.; (2) then we construct the feedback matrices $\mat{B}^l$ with spectral norm (largest singular value) equal to $\beta_l$ (an algorithm parameter); the entries of $\mat{B}^l$ are sampled independently from the standard Gaussian distribution, and then this matrix is rescaled by a constant so that $||\mat{B}^l||=\beta_l$; (3) we construct a mini-batch of size $m$ by sampling without replacement; (4) after the feed-forward phase completes, we compute the clipped version (with parameter $\tau_h$) of the post-activation $\vec{h}_l$ of each layer  and we also clip with parameter $\tau_e$ the error vector $\vec{e}$; (5) we then add Gaussian noise to the DFA mini-batch update direction. The full algorithm is shown in Algorithm \ref{alg:dp-dfa}.

\IncMargin{1em}
\begin{algorithm}[tp]
  \DontPrintSemicolon
  \KwIn{A feed-forward network $f_{\theta}=\{(\mat{W}^l,
    \vec{b}^l)\}_{l=1}^L$, upper bounds $\gamma_l$ on derivative of activation functions, spectral norm bounds $\beta_l$, clipping
    thresholds $\tau_e, \tau_h$, stepsize $\eta$, noise scale $\sigma$, minibatch size $m$, number of iterations $T$.}
  \For{$l=1$ \KwTo $L$}{%
    Intialize $\mat{B}^l$,\, then normalize:
    $\mat{B}^l = \beta_l \mat{B}^l / \norm{\mat{B}^l}_2$ \tcp*{spectral normalization}
  }
  \For{$T$ iterations}{%
    Create a minibatch of size $m$ by sampling without replacement\;
    $S\gets$ indices of records in the mini-batch\;
    \For{$i$ in $S$} {
    \For(\tcp*[h]{feed-forward phase}){layer $l=1$ \KwTo $L-1$}{%
      $\vec{z}^l_i = \mat{W}^l\vec{h}_i^{l-1}+\vec{b}^l$;\,
      $\vec{h}_i^l = \phi_l(\vec{z}_i^l)$;\,
      \tcp*{process record $i$}
    }
    $\hat{\vec{y}}_i \gets \mathsf{softmax}(\mat{W}^L\vec{h}_i^{L-1}+\vec{b}^L)$\;
    $\vec{e}_i = \mathsf{clip}_{\tau_e}(\hat{\vec{y}}_i -\vec{y}_i)$
    \tcp*{error clipping}
    }
    \For(\tcp*[h]{activation clipping and parameter updates}){$l=L$ \KwTo $1$}{%
      $\mat{W}^l \gets \mat{W}^l -
      \eta\left(\mathcal{N}(0,
        \sigma^2\mat{I}) + \frac{1}{m}\sum_{i\in S}\left((\mat{B}^{l+1}\vec{e}_i)\odot
          \phi'_l(\vec{z}_i^l)\right)\mathsf{clip}_{\tau_h}(\vec{h}_i^{l-1})^\intercal  \right)$\label{line:w}\; 
      $ \vec{b}^l \gets \vec{b}^l - \eta\left(\mathcal{N}(0,
        \sigma^2\mat{I}) + \frac{1}{m}\sum_{i\in S}\left(\mat{B}^{l+1}\vec{e}_i\right) \odot \phi_l'(\vec{z}_i^l)\right)\label{line:b}\;$
    }
  }
  \caption{DP-DFA with activation clipping}
  \label{alg:dp-dfa}
\end{algorithm}
\DecMargin{1em}

\subsection{Privacy Accounting}\label{subsec:account}
At the end of the day, a user is interested in computing the RDP parameters of Algorithm \ref{alg:dp-dfa} and then converting it into the $\epsilon,\delta$ parameters of differential privacy. We now describe this process. The $T$ iterations of the algorithm correspond to the sequential composition of $T$ mechanims $M_1,\dots, M_T$, where $M_j$ applies the Gaussian Mechanism (lines \ref{line:w} and \ref{line:b}) on sampled data (the minibatch) of iteration $j$ of the algorithm. Let $\Delta$ be the sensitivity of the combined computations (for all layers combined) of $\frac{1}{m}\sum_{i\in S}\left((\mat{B}^{l+1}\vec{e}_i)\odot
          \phi'_l(\vec{z}_i^l)\right)\mathsf{clip}_{\tau_h}(\vec{h}_i^{l-1})^\intercal$ and $ \frac{1}{m}\sum_{i\in S}\left(\mat{B}^{l+1}\vec{e}_i\right) \odot \phi'_l(\vec{z}_i^l)$ (we show how to compute this in Section \ref{subsec:sen}). The Gaussian noise (Lemma \ref{lemma:gaussmech}) added to these quantities provides $(\alpha, \alpha\Delta^2/(2\sigma^2))$-RDP. The second RDP parameter is further reduced by applying Lemma \ref{lemma:rdp_amp} since the batch was chosen randomly. The second parameter is then multiplied by $T$ to account for all iterations. Finally, the resulting RDP parameters are converted to $(\epsilon, \delta)$-differential privacy parameters using Lemma \ref{pro:conversion}.

\subsection{Sensitivity Computation}\label{subsec:sen}
Thus, all that is left is to compute the sensitivity $\Delta_{\zeta_l}$ of $\zeta_l\equiv \frac{1}{m}\sum_{i\in S}\left((\mat{B}^{l+1}\vec{e}_i)\odot
          \phi'_l(\vec{z}_i^l)\right)\mathsf{clip}_{\tau_h}(\vec{h}_i^{l-1})^\intercal$ and the sensitivity $\Delta_{\xi_l}$ of  $\xi_l\equiv \frac{1}{m}\sum_{i\in S}\left(\mat{B}^{l+1}\vec{e}_i\right) \odot \phi'_l(\vec{z}_i^l)$ under the "modify one record" version of neighboring datasets (Secction \ref{subsec:dp}). 
          
Once we have those quantities, the overall  sensitivity $\Delta$ (used in Section \ref{subsec:account}) is clearly equal to $(L\Delta_{\zeta_l}^2 +  L\Delta_{\xi_l}^2)^{1/2}$.

\noindent\textbf{Sensitivity for Fully Connected Networks}. For fully connected networks, we note that changing a record only changes one term in the summations, so the sensitivity of $\zeta_\ell$ is equal to
\begin{align*}
    \Delta_{\zeta_l}\leq \frac{2}{m}\norm*{\left((\mat{B}^{l+1}\vec{e}_i)\odot
          \phi'_l(\vec{z}_i^l)\right)\mathsf{clip}_{\tau_h}(\vec{h}_i^{l-1})^\intercal}_F &= \frac{2}{m}\norm*{\left(\mat{B}^{l+1}\vec{e}_i\right) \odot \phi_l'(\vec{z}_i^l)}_2 
    \norm{\mathsf{clip}_{\tau_h}(\vec{h}_i^{l-1})}_2\\
    &\leq \frac{2}{m}\gamma_l \tau_h \norm*{\mat{B}^{l+1}\vec{e}_i}_2 = \frac{2}{m}\gamma_l \tau_h \beta_{l+1}\tau_e
\end{align*}
while for $\xi_\ell$, a similar computation for sensitivity yields:
\begin{align*}
    \Delta_{\xi_l}\leq \frac{2}{m}\norm*{(\mat{B}^{l+1}\vec{e}_i)\odot
          \phi'_l(\vec{z}_i^l)}_2 
    \leq \frac{2}{m}\gamma_l  \norm*{\mat{B}^{l+1}\vec{e}_i}_2 = \frac{2}{m}\gamma_l   \beta_{l+1}\tau_e
\end{align*}
So the overall sensitivity of one iteration, to be used in the privacy accounting in Section \ref{subsec:account}, is $\Delta \leq 2\gamma_l \beta_{l+1}\tau_e \sqrt{(1+\tau_h^2)L}/m$.

\noindent\textbf{Handling Convolutional Layers}. Although DFA can be extended to convolutional layers,  several results suggest that it is better to use a hybrid BP/DFA approach instead. The first reason is that in the non-private setting, it has been observed that DFA does not perform well with convolutional layers \cite{Han_2019_ICCV}. Second, we observed that this rule sometimes carries over to the privacy-preserving setting. The sensitivity is much larger than for fully connected layers.\footnote{For completeness, we include the details in the supplementary material.} 

In the non-private setting, Han et al. \cite{Han_2019_ICCV} suggested a hybrid approach that we can carry over to the privacy-preserving setting. This hybrid approach can be visualized in Figure \ref{fig:CNN_hybrid}. Consider a network with layers $1,\dots, \ell$ being convolutional and layers $\ell+1, \dots, L$ being fully connected.

 For the fully connected layers (i.e., $\ell+1,\dots, L$), we use the differentially private DFA updates from lines \ref{line:w} and \ref{line:b} from Algorithm \ref{alg:dp-dfa}. Given an overall target sensitivity $C$, we choose the algorithm parameters so that the sensitivity of each fully connected layer is $C/\sqrt{L}$.

 For the convolutional layers (i.e., $1,\dots, \ell$), the actual gradient at layer $l$ is  $\pdv{\mathcal{L}}{\mat{W}^l}
  = \pdv{\vec{z}^{\ell+1}}{\mat{W}^l}\pdv{L}{\vec{z}^{\ell+1}}$. However, since we did not back-propagate from $L$ down to $\ell+1$, the partial derivative $\pdv{L}{\vec{z}^{\ell+1}}$ is not available by the time the algorithm is processing the convolutional layers. Hence we replace $\pdv{L}{\vec{z}^{\ell+1}}$ with its DFA counterpart $\delta\vec{z}^{\ell+1}= \left(\mat{B}^{\ell+2}\vec{e}\right) \odot \phi_{\ell+1}'(\vec{z}^{\ell+1})$ (see Equation \ref{eq:dfa_error_prop}). Meanwhile,  $\pdv{\vec{z}^{\ell+1}}{\mat{W}^l}$ can be computed using back-propagation starting at layer $\ell+1$. Putting this together, the differentially private update step for convolutional layers that should be performed on lines \ref{line:w} and \ref{line:b} is:
  \begin{align*}
       \mat{W}^l &\gets \mat{W}^l -
      \eta\left(\mathcal{N}(0,
        \sigma^2\mat{I}) + \frac{1}{m}\sum_{i\in S}\mathsf{clip}_\tau\left(\pdv{\vec{z}^{\ell+1}_i}{\mat{W}^l}  \delta\vec{z}_i^{\ell+1}\right)\right)\\
       \vec{b}^l &\gets \vec{b}^l - \eta\left(\mathcal{N}(0,
        \sigma^2\mat{I}) + \frac{1}{m}\sum_{i\in S}\mathsf{clip}_\tau\left(\pdv{\vec{z}^{\ell+1}_i}{\vec{b}^l}  \delta\vec{z}_i^{\ell+1}\right)\right)
  \end{align*}
  where the clipping threshold $\tau$ is chosen so that the sensitivity of each convolutional layer is $C/\sqrt{L}$.
  
  Since each network layer has sensitivity $C/\sqrt{L}$, their combined $L_2$ sensitivity is $(C^2/L + \cdots + C^2/L)^{1/2}=C$.



\section{Exerimental Results}
\label{sec:experiments}
In this section, we compare differentially private DFA (DP DFA) to differentially private BP (DP BP) \cite{Abadi2016deep}.
It is first worth mentioning some struggles with DP BP reported in prior work (e.g., \cite{Abadi2016deep,bagdasaryan2019differential}). Earlier work used pre-trained convolutional layers that were never updated \cite{Abadi2016deep} so that those experiments only performed DP BP over 2 or 3 fully connected layers. Subsequent work (e.g., \cite{bagdasaryan2019differential}) used more complex networks (Inception V3) that were pretrained; however, DP BP caused their accuracy to decrease almost immediately. For these reasons, we use networks in similar complexity to \cite{Abadi2016deep}. However, we train \textbf{all} the layers in our networks and furthermore show that the models can actually be trained from scratch.\footnote{This eliminates the temptation of practitioners to misuse differential privacy by pre-training on private data and only afterwards using differentially private updates.} 

We evaluate performance on Fashion MNIST
(FMNIST~\cite{xiao2017/online}) and
CIFAR10~\cite{Krizhevsky09learningmultiple}. In all cases, the differentially private update directions are sent to the Adam optimizer \cite{Kingma2014adam} (with parameters $\eta=0.001, \beta_1=0.9, \beta_2=0.999$). In all experiments, the $\beta$ parameter for DP DFA was set to 0.9 and the error clipping threshold was $\tau_e=0.1$. The activation clipping threshold $\tau_h$ was then set so that the overall sensitivity matched that of DP BP.

\begin{figure}[tp]
  \centering
  \begin{subfigure}[b]{.3\textwidth}
    \includegraphics[width=\textwidth]{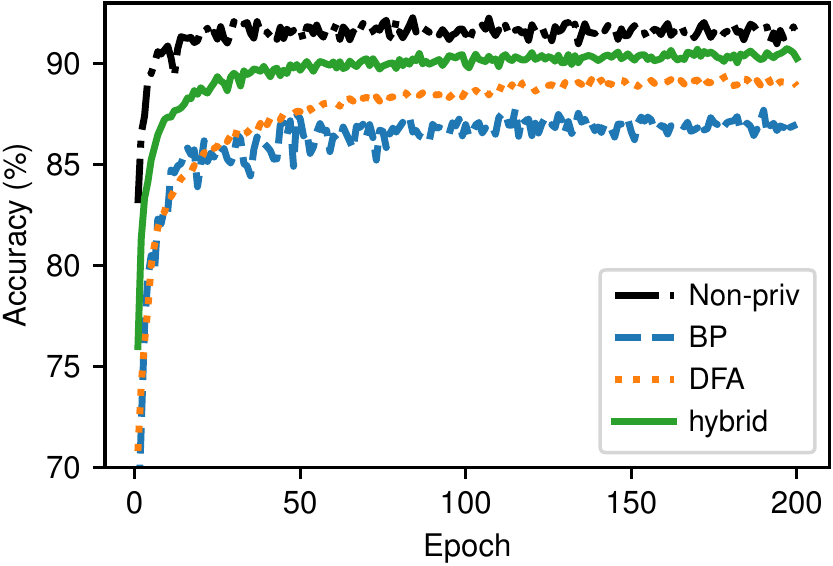}
    \caption{Test Accuracy, $\sigma=0.01$. At $\delta=10^{-5}$, the
      $\epsilon$ values at epochs 50, 100, 150, 200 are
      4.43, 6.43, 8.07, and 9.49, respectively.}
  \end{subfigure}$\quad$
  \begin{subfigure}[b]{.3\textwidth}
    \includegraphics[width=\textwidth]{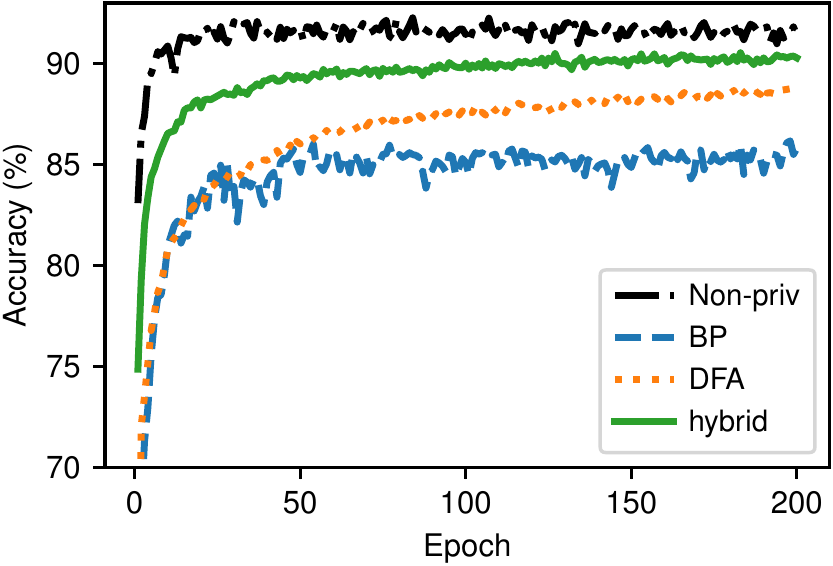}
    \caption{Test Accuracy, $\sigma=0.03$. At $\delta=10^{-5}$, the
      $\epsilon$ values at epochs 50, 100, 150, 200 are
      1.28, 1.82, 2.25, and 2.61, respectively.}
  \end{subfigure}  $\quad$
  \begin{subfigure}[b]{.3\textwidth}
    \includegraphics[width=\textwidth]{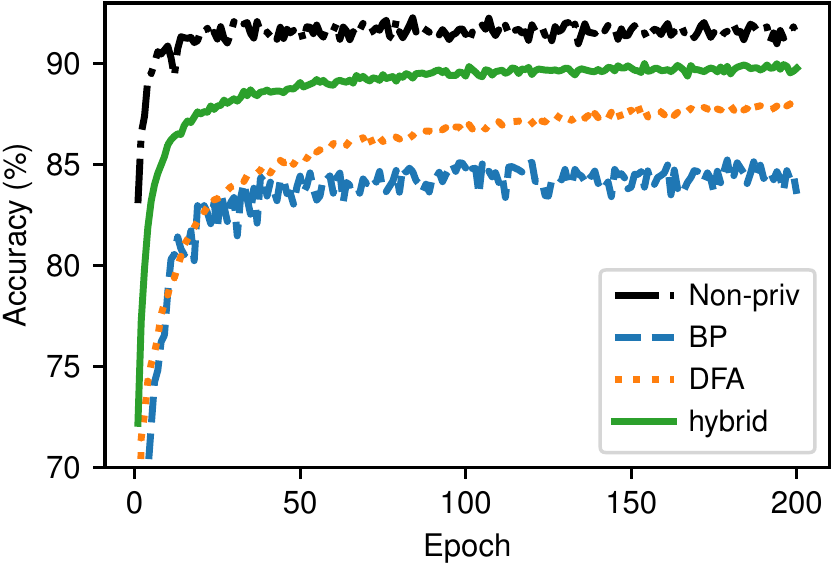}
    \caption{Test Accuracy, $\sigma=0.05$. At $\delta=10^{-5}$, the
      $\epsilon$ values at epochs 50, 100, 150, 200 are
      0.75, 1.07, 1.31, and 1.52, respectively.} 
  \end{subfigure}
  \caption{Differentially private convolutional network experiments on Fashion-MNIST. Test Accuracy is reported after every training epoch.}\label{fig:exp_conv_fmnist}
\end{figure}

\begin{figure}[tp]
  \centering
  \begin{subfigure}[b]{.3\textwidth}
    \includegraphics[width=\textwidth]{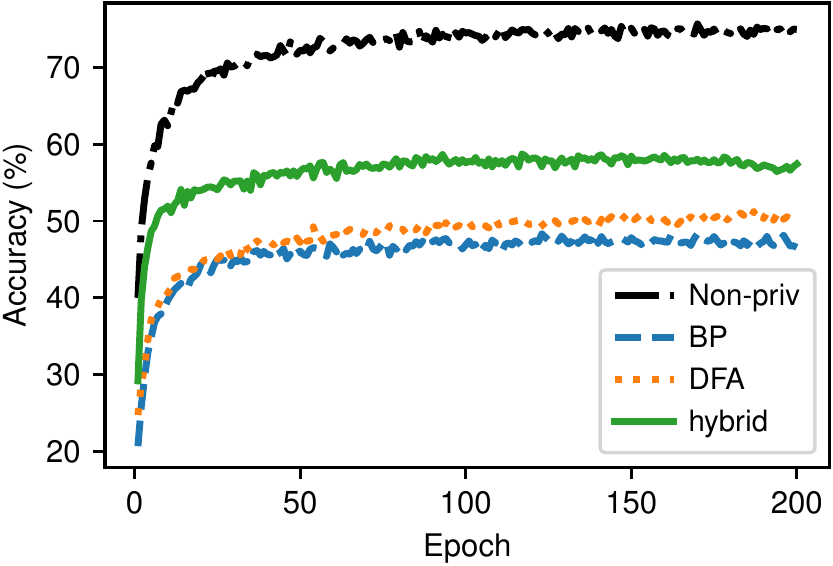}
    \caption{Test Accuracy $\sigma=0.01$}
  \end{subfigure}$\quad$
  \begin{subfigure}[b]{.3\textwidth}
    \includegraphics[width=\textwidth]{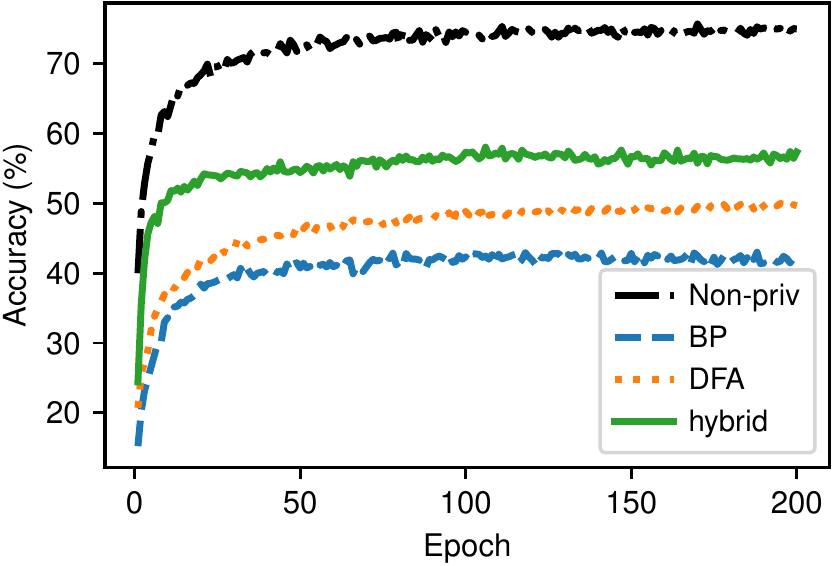}
    \caption{Test Accuracy $\sigma=0.03$}
  \end{subfigure}$\quad$  
  \begin{subfigure}[b]{.3\textwidth}
    \includegraphics[width=\textwidth]{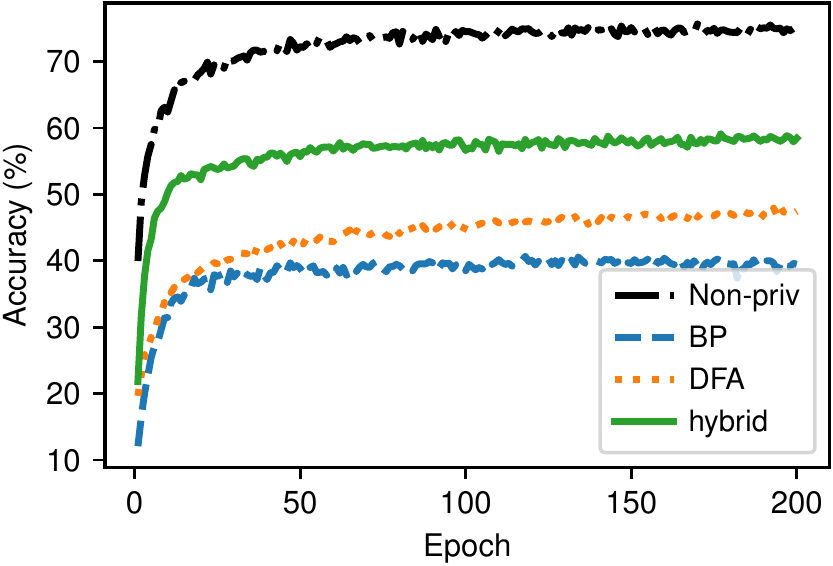}
    \caption{Test Accuracy $\sigma=0.05$}
  \end{subfigure}
\caption{Differentially private convolutional network experiments on CIFAR 10. Test Accuracy is reported after every training epoch. Privacy parameters at different epochs are the same as in Figure \ref{fig:exp_conv_fmnist}}\label{fig:exp_conv_cifar10}
\end{figure}

\subsection{Convolutional Networks.}
For convolutional networks, we use a similar setup to \cite{Abadi2016deep}. The architecture uses 2 convolutional layers (5x5 kernels and 64 channels each with 2x2 max pools) followed by 2 fully connected layers (384 units each), and the output layer uses softmax. Following \cite{Abadi2016deep}, we set the sensitivity for each iteration at 3 (so even the DP DFA uses their setting) and a mini-batch of size 512. For DP BP, following \cite{Abadi2016deep}, the layers use ReLU activation. ReLU is  not recommended for DFA in the non-private case \cite{nokland2016direct} so for DP DFA we use tanh for convolutional layers and sigmoid for fully connected layers (we use this setting even when using the differentially private DFA/BP hybrid).

This setup ensures that each iteration and each epoch of DP BP has the same privacy impact as DP DFA. The results for Fashion MNIST are shown in Figure \ref{fig:exp_conv_fmnist}. At various noise levels $\sigma$, we see that a large gap between differentially private BP and the non-private network. Differentially private DFA outperforms DP BP, with the hybrid approach clearly outperforming both, and doing a better job of closing the gap with the non-private network. Corresponding results for CIFAR 10, which is known to be much harder for differentially private training, is shown in Figure \ref{fig:exp_conv_cifar10}. We see the same qualitative results, where the differentially private hybrid approach significantly outperforms the other privacy-preserving methods.

\subsection{Fully Connected Networks.}

\begin{figure}[tp]
  \centering
  \begin{subfigure}[b]{.9\textwidth}
    \centering
    \includegraphics[width=.35\textwidth]{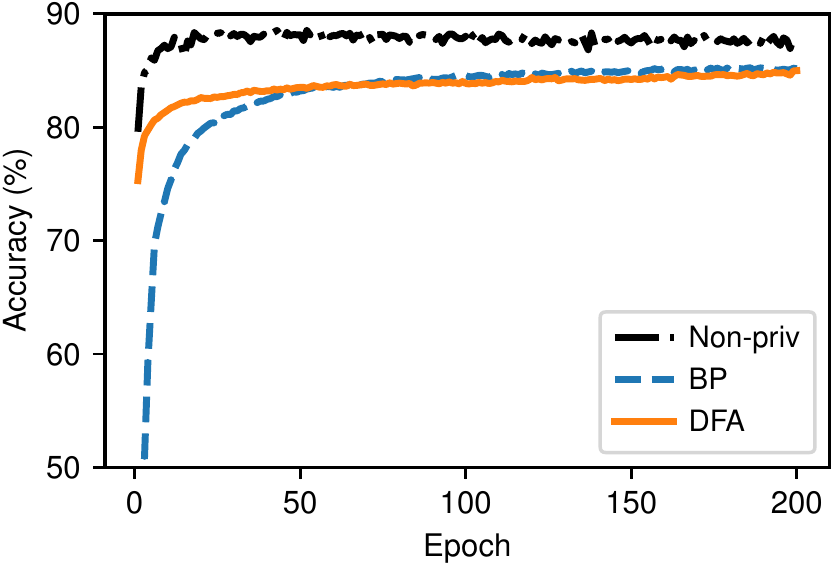}
    \includegraphics[width=.35\textwidth]{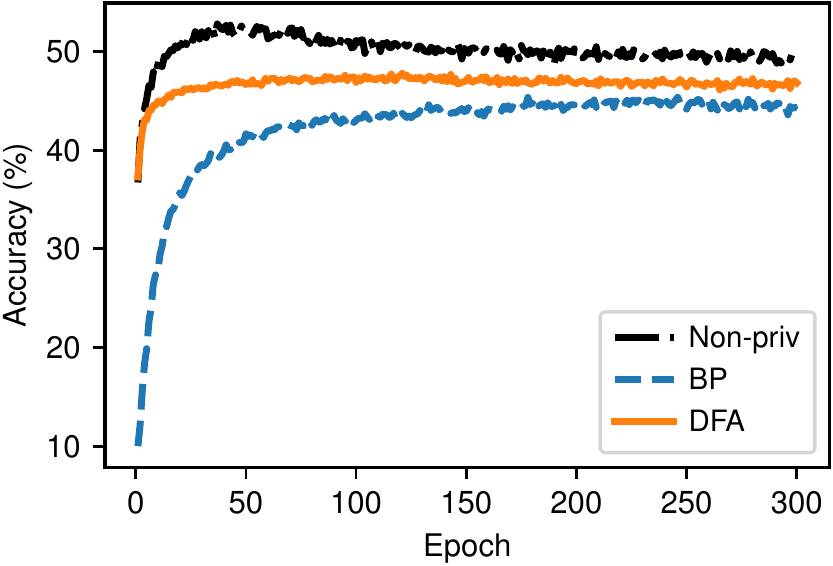}
    \caption{Testing accuracy, $\sigma=0.01$. At $\delta=10^{-5}$, the
      $\epsilon$ values at epochs 50, 100, 150, 200, 250, 300 are 9.77,
      13.75, 17.8, 21.3, 23.74, and 26.19, respectively. (left: Fashon-MNIST,
      right CIFAR10)}
  \end{subfigure}~\\
  \begin{subfigure}[b]{.9\textwidth}
    \centering
    \includegraphics[width=.35\textwidth]{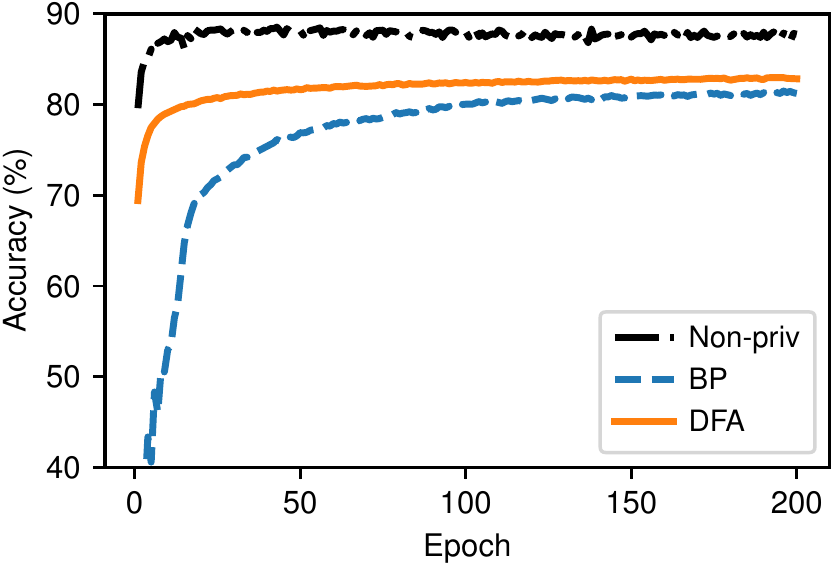}
    \includegraphics[width=.35\textwidth]{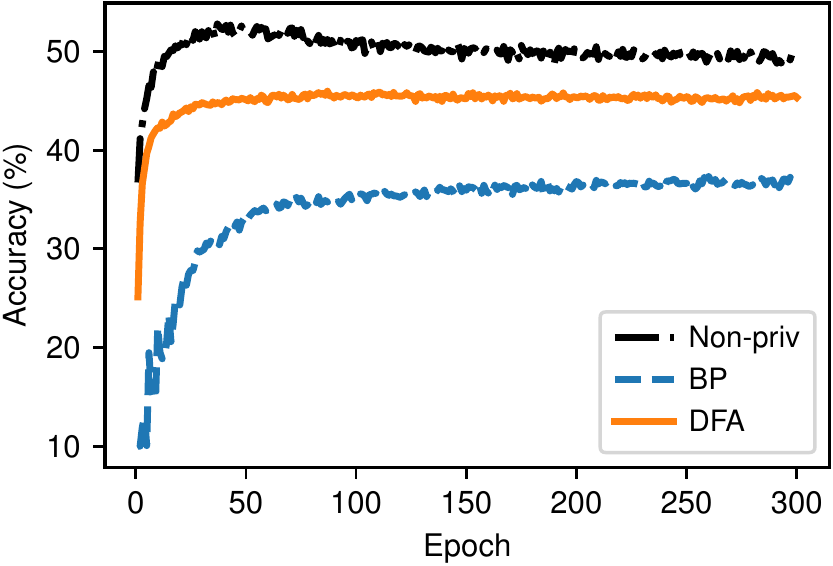}
    \caption{Testing accuracy $\sigma=0.05$. At $\delta=10^{-5}$,
      the $\epsilon$ values at epochs 50, 100, 150, 200, 250, 300 are
      1.03, 1.46, 1.8,2.1,2.35, and 2.59, respectively. (left: Fashon-MNIST,
    right CIFAR10)}
  \end{subfigure}
  \caption{Differentially private fully connected network  on Fashion-MNIST and CIFAR 10}
  \label{fig:connected}
\end{figure}

For fully connected networks, we use the following architectures. For Fashion MNIST,  the network consists of two hidden layers, with 128 and 256 hidden units, respectively. These layers use the sigmoid activation function and we use softmax for the output layer. We use a minibatch of size 128 and the sensitivity of each iteration is set to 2. As usual, the privacy impact of each iteration and epoch is the same for DP BP and for DP DFA. Since CIFAR 10 is a more complex dataset, we use a more complex network with 3 hidden layers of 256 units each.  The results are shown in Figure \ref{fig:connected}. We see that the accuracy of all networks saturate fairly quickly with DP DFA achieving good accuracy with fewer epochs than DP BP, suggesting that we can use it to train more accurate networks with a smaller privacy budget.

\section{Conclusion}
\label{sec:conclusion}
Recent advances in differential privacy have shown that privacy-preserving training of complex non-convex models is feasible. However, there is still a significant gap between accuracy of these models and application requirements. In this paper, we consider the possibility that alternatives to backprop may be more suitable for training privacy-preserving models. We proposed the first differentially private version of direct feedback alignment (DFA), a biologically inspired training algorithm. 

Although the effects of DFA are not well-understood even in the non-private setting, its behavior in the privacy-preserving setting (when compared to differentially private SGD) and potential use in low-power hardware show that DFA merits increased attention and theoretical study.





\small
\bibliographystyle{abbrv}
\bibliography{reference}


\end{document}